\documentclass[10pt,journal,compsoc]{IEEEtran}

\usepackage{amsmath,amsfonts}
\usepackage{array}
\usepackage[caption=false,font=normalsize,labelfont=sf,textfont=sf]{subfig}
\usepackage{textcomp}
\usepackage{stfloats}
\usepackage{url}
\usepackage{verbatim}
\usepackage{graphicx}
\usepackage{cite}
\usepackage{times}
\usepackage{epsfig}
\usepackage[dvipsnames]{xcolor}
\usepackage{amssymb}
\usepackage{multirow}
\usepackage{makecell}
\usepackage{listings}
\usepackage{algorithm}
\usepackage{algorithmicx}
\usepackage{booktabs}
\usepackage{hyperref}

\definecolor{dkgreen}{rgb}{0,0.6,0}
\definecolor{gray}{rgb}{0.5,0.5,0.5}
\definecolor{mauve}{rgb}{0.58,0,0.82}

\begin{document}

\title{SpikeCV: Open a Continuous Computer\\Vision Era}

\author{\IEEEauthorblockN{Yajing Zheng, Jiyuan Zhang, Rui Zhao, Jianhao Ding, Shiyan Chen, \\ Ruiqin Xiong, Zhaofei Yu, and Tiejun Huang}
\thanks{Y. Zheng, J. Zhang, R. Zhao, J. Ding and S. Chen are with the School of Computer Science, Peking University, Beijing, China. E-mail:yj.zheng@pku.edu.cn, \{jyzhang, ruizhao, djh01998, 2001212818\}@stu.pku.edu.cn.
}
\thanks{R. Xiong is with the School of Computer Science, Peking University, Beijing, China. E-mail: rqxiong@pku.edu.cn.}
\thanks{Z. Yu and T. Huang are with the Institute for Artificial Intelligence, Peking University, Beijing, China, and also with school of Computer Science, Peking University, Beijing, China. E-mail:\{yuzf12, tjhuang\}@pku.edu.cn}
}




\IEEEtitleabstractindextext{
\begin{abstract}
   SpikeCV is a new open-source computer vision platform for the spike camera, which is a neuromorphic visual sensor that has developed rapidly in recent years. In the spike camera, each pixel position directly accumulates the light intensity and asynchronously fires spikes. The output binary spikes can reach a frequency of 40,000 Hz. As a new type of visual expression, spike sequence has high spatiotemporal completeness and preserves the continuous visual information of the external world. Taking advantage of the low latency and high dynamic range of the spike camera, many spike-based algorithms have made significant progress, such as high-quality imaging and ultra-high-speed target detection. 
   To build up a community ecology for the spike vision to facilitate more users to take advantage of the spike camera, SpikeCV provides a variety of ultra-high-speed scene datasets, hardware interfaces, and an easy-to-use modules library. SpikeCV focuses on encapsulation for spike data, standardization for dataset interfaces, modularization for vision tasks, and real-time applications for challenging scenes. With the advent of the open-source Python ecosystem, modules of SpikeCV can be used as a Python library to fulfilled most of the numerical analysis needs of researchers. We demonstrate the efficiency of the SpikeCV on offline inference and real-time applications. The project repository address are \url{https://openi.pcl.ac.cn/Cordium/SpikeCV} and \url{https://github.com/Zyj061/SpikeCV}.
\end{abstract}

}

\maketitle

\IEEEdisplaynontitleabstractindextext

\IEEEpeerreviewmaketitle

\section{Introduction}

Traditional digital cameras use a continuous sequence of frames to represent changes in light signals. Although the shooting process is continuous, there is a fixed exposure time interval between two consecutive frames. Visual changes during the interval are thus lost. For changing scenes, recording continuous vision should pay more attention to the light signal itself. It is written in The Computers and the Brain~\cite{von2012computer} by von Neumann that discrete nerve spikes can express continuous analog values, such as light, sounds, etc. For expressing continuous vision, the spike camera~\cite{huang2022spiking,huang20221000x} imitates the working mechanism of the retina, representing photons intensity process with a continuous values sequence—intervals of the spike train.

Collecting continuous visual information has great advantages in analyzing challenging scenes, such as analyzing the instantaneous state of high-speed moving molecules/objects~\cite{uchihashi2012guide}, monitoring high-speed machining processes in the industry~\cite{haber2004investigation}, or rapid obstacle avoidance in drones or automatic driving~\cite{falanga2020dynamic}. In recent years, taking advantage of the low latency and high dynamic range (HDR) of the spike camera, many spike-based works have emerged, including high-quality image reconstruction~\cite{zhao2020high,zhao2021spk2imgnet,zheng2021high,zheng2023capture,zhu2019retina}, optical flow~\cite{hu2022optical,zhao2022spikingsim}, depth estimation~\cite{wang2022learning,zhang2022spike}, and ultra-high-speed target detection and tracking~\cite{huang20221000x,zheng2022spike}, and HDR imaging~\cite{han2023hybrid,han2020neuromorphic,zhou2020unmodnet}. In addition to applying spike cameras to vision tasks, there are works dedicated to coding and simulators for spike cameras~\cite{hu2022optical,zhao2022spikingsim}. 

\begin{figure}
    \centering
    \includegraphics[width=0.98\linewidth]{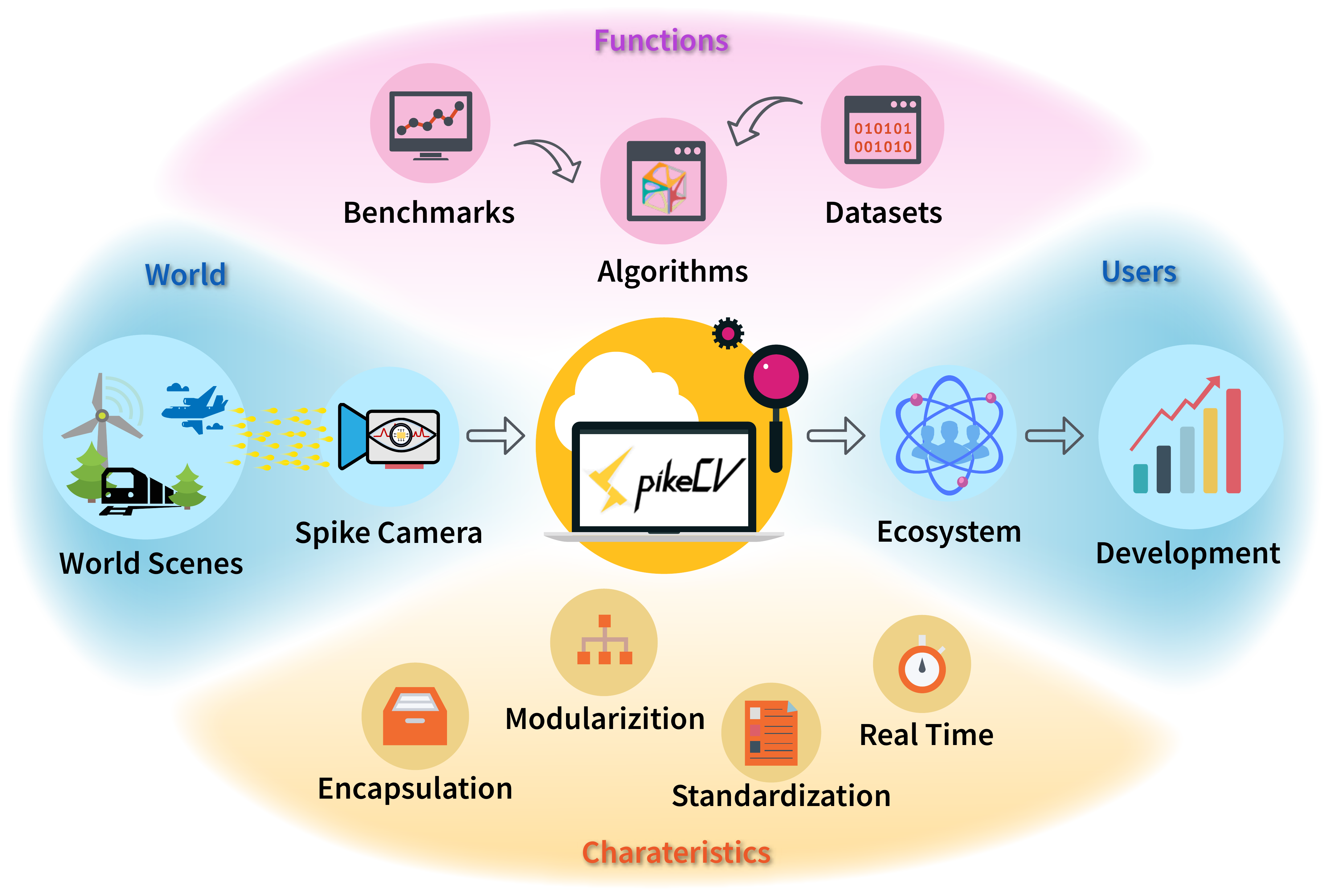}
    \caption{Macro design principles of SpikeCV. With the functions and characteristics of SpikeCV, we can obtain continuous visual information of the world through the spike camera, and promote the development of the spike vision ecosystem.}
    \label{fig:head-figure}
\end{figure}

Unlike the sequence of frames employed in traditional vision, the output of the spike camera is a stream of bits representing light signals changing at ultra-fast speed. However, almost all tools in computer vision serve image sequences, such as OpenCV~\cite{bradski2008learning}.
As a Chinese idiom says, \textit{to do a good job, an artisan needs the best tools}. At present, although the spike camera and the spike-based algorithms have made some staged progress, it is still hard for novices to get started with spike vision. Therefore, to better promote the advancement of machine vision using continuous vision based on spikes, we build a spike-based open-source platform for spike camera, \textit{SpikeCV}.

Unlike existing open-source frameworks that mainly provide image processing or algorithm tools, SpikeCV provides spike processing tools and spike-based algorithms, hardware interfaces of spike cameras, and standardized spike datasets. To facilitate users to program and use numerical analysis toolkits (e.g., NumPy~\cite{oliphant2006guide}, SciPy~\cite{virtanen2020scipy} and PyTorch~\cite{paszke2019pytorch}), SpikeCV mainly uses Python as the programming language, and C++ for real-time interacting with the hardware. An intuitive illustration of the design principles of SpikeCV is shown in Fig.~\ref{fig:head-figure}. SpikeCV is a bridge between spike cameras and users. By providing different functional modules and easy-to-use packaging interfaces, standardized data, real-time processing pipelines, and modular spike-based tools, it helps to form a spike vision community and promote the development of continuous vision scene applications. In this paper, we introduce the design principles, working mechanism, and usage of SpikeCV in detail. In the experiments, we demonstrate the efficiency of SpikeCV by achieving multiple vision tasks in the same scene offline and in real time, respectively. Our contributions can be summarized as:

\begin{itemize}
\setlength{\itemsep}{0pt}
\setlength{\parsep}{0pt}
\setlength{\parskip}{0pt}
    \item We provide the first open-source platform for spike cameras, encapsulate datasets, device interface, spike loader, and spike-based algorithms, to make it easy to build spike vision applications.
    \item We standardize the datasets of various visual tasks, and provide a unified data interface for different visual tasks, offline files, and real-time spike streams. 
    \item We realize the real-time pipeline for hardware reading of various spike cameras by providing hardware-friendly C++ and user-friendly Python interface, which achieve the synchronization between reading and processing of spike data.
\end{itemize}




\section{Background}
\subsection{Mechanism of Spike Camera} 
\begin{figure}
    \centering
    \includegraphics[width=0.98\linewidth]{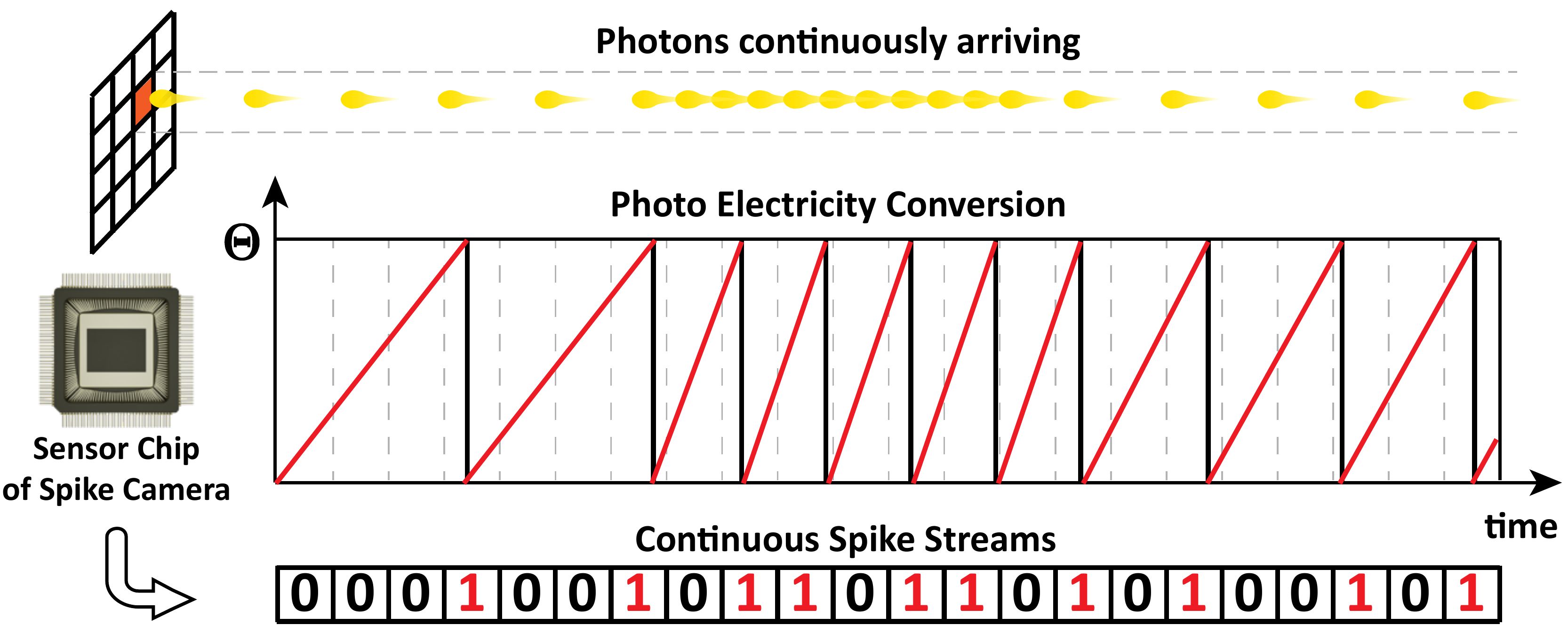}
    \caption{Mechanism of spike camera.}
    \label{fig:mechanism_camera}
\end{figure}

The sensor of a spike camera is arranged by $H \times W$ pixel units that continuously receive photons and fire spikes~\cite{dong2017spike}. As illustrated in Fig.~\ref{fig:mechanism_camera}, the instant light intensity $L_{i,j}(t)$ at position $(i,j)$ will be converted to an electrical signal to increase voltage $V_{i,j}(t)$ at time $t$. The voltage $V_{i,j}(t)$ will be reset to $0$ when it reaches a threshold $\Theta$, at which a spike will be fired. The process can be formulated as:
\begin{align}
    \label{spk1}
    V_{i,j}(t) = \left( \int_{t^{\text{last}}_{i,j}}^{t} \alpha\cdot L_{i,j}(\phi) d \phi  \right)\bmod{\Theta},
\end{align}
where $t^{\text{last}}_{i,j}$ is the time when the last spike fired at $(i,j)$. The back-end synchronous circuit always poll the $V(t)$ with a very small interval $\tau$($25\mu s$) and generate a binary spike stream $S$:
\begin{align}
    \label{spk2}
    S_{i,j,k} =  \left\{
    \begin{array}{c}
        1, \space \text{if } \space \exists t \in ((k-1)\tau,k\tau], V_{i,j}(t)=0,\\
        0, \space \text{otherwise} \quad\quad\quad\quad\quad\quad\quad\quad\quad\quad\quad \\
    \end{array}
    \right.
\end{align}
where $k=1,2,...$ is the $k$-th polling. Thus, for $T$ times polling, the output $S$ with a size of $H \times W \times T$ will be generated.

\subsection{Spike-based Algorithms}

The research works related to spike cameras have made significant progress, which can be divided into three categories: Coding and simulation of spikes, High-quality imaging based on spikes, and Vision tasks based on spikes. These three parts correspond to the bottom-up research of the spike vision.

\subsubsection{Coding and simulation of spikes.} Spike coding is mainly to compress the dense binary spike streams through statistical characteristics, which is similar to the field of traditional video coding and decoding~\cite{girod2005distributed, ohm2012comparison}. Spike coding methods include lossy~\cite{zhu2020hybrid} and lossless compression~\cite{dong2019efficient}. The research of spike simulation~\cite{hu2022optical,zhao2022spikingsim} is mainly to explore and improve the working mechanism of spike cameras and obtain simulated spike data under many extreme conditions, which is convenient for the development of spike vision research. Deep learning has been successfully applied to various vision/natural language tasks~\cite{lecun2015deep,ramachandram2017deep,vaswani2017attention}, but its training requires a large amount of training data. Acquiring large-scale datasets for ultra-high temporal resolution spike arrays is costly, and the spike simulator can generate large-scale spike data with ground truths, such as high-frequency images, optical flow, and depth. Zhao et al.~\cite{zhao2022spikingsim} proposed the SpikingSIM simulator for generating spike data from images based on the sensor principle of the spike camera. Hu et al.~\cite{hu2022optical} proposed SPCS to generate spike data and labels based on graphics models~\cite{blender}.

\subsubsection{High-quality imaging based on spikes.} Spike cameras preserve the high-resolution spatiotemporal information of continuous vision, making them ideal for image reconstruction~\cite{zhao2021spk2imgnet,zheng2021high} and high-dynamic-range imaging~\cite{han2023hybrid,han2020neuromorphic,zhou2020unmodnet}. The current high-quality image reconstruction algorithms for spike cameras can be divided into statistics-based~\cite{zhu2019retina, zhao2020high}, deep learning-based methods~\cite{zhao2021spk2imgnet}, and spiking neural networks (SNNs)-based~\cite{bi1998synaptic,maass1997networks,tsodyks1997neural} methods~\cite{zheng2021high, zheng2023capture,zhu2020retina}. The main principle of the Statistical method is that the spike emission frequency is proportional to the light intensity, and the pixel value can be deduced according to the interval or the frequency of spikes in the sliding window. Deep learning-based reconstruction algorithms include supervised learning networks~\cite{wu1998piecewise} using simulated data~\cite{zhao2021spk2imgnet, zhao2021reconstructing}, and self-supervised reconstruction algorithms~\cite{chen2022self} based on blind spot networks~\cite{ lehtinen2018noise2noise}. In addition to high-speed and high-dynamic-range imaging, some recent research~\cite{xiang2021learning,zhao2021super} focus on obtaining super-resolved images from spike data.

\subsubsection{Vision tasks based on spikes.} Besides reconstructing images, there are some vision algorithms methods using spatiotemporal information directly from spike data, including networks for optical flow~\cite{hu2022optical, zhao2022learning}, depth~\cite{wang2022learning,zhang2022spike}, detection and tracking~\cite{huang20221000x,li2022retinomorphic,zhu2022fpga}. 
SCFlow~\cite{hu2022optical} is a deep pyramidal network with a motion-adaptive representation for spike-based optical flow, trained by graphics-based simulated data. Spike2Flow~\cite{zhao2022learning} explores the continuousness of spike data in optical flow. Wang et al.~\cite{wang2022learning} propose the first stereo depth network for spike data with Transformers based on self and cross-attention. Zhang et al.~\cite{zhang2022spike} propose to use a 3D Swin Transformer to represent spike stream and a spatial-temporal Transformer to estimate monocular depth. Huang et al.~\cite{huang20221000x}, and Zheng et al. ~\cite{zheng2022spike} propose to use SNNs to implement high-speed detection and tracking through spike cameras. Zhu et al.~\cite{zhu2022fpga} use FPGA to improve the speed of spike-based detection and tracking. Li et al.~\cite{li2022retinomorphic} propose to use both spike camera and event camera to imitate the retina for object detection.
 

\section{Design Principles}

\subsection{Usability}
\subsubsection{Encapsulation for Spikes}
\label{sec:encapsu_spikes}
At present, the spike camera has released its second generation, and there would be new-coming ones with extra characteristics whose spatial resolution, sampling rate, and data channels may be various. Besides, implementations and representations of spikes vary from different algorithms and different visual tasks. Therefore, with the development of spike cameras, the community requires a unified interface to represent such a new modality. In the SpikeCV, we design the new class \textbf{SpikeStream} to encapsulate the spike data.

\begin{figure}[htb]
\rule{0.47\textwidth}{1pt}
\begin{minipage}{0.45\textwidth}
\begin{lstlisting}
class SpikeStream:
    def __init__(self, offline=True, **kwargs):
        self.SpikeMatrix = None
        self.offline = offline
        # assign parameters according to the type of input stream
        if self.offline: 
            # read offline data from file
            self.filename, self.height, self.width = ParseKwargs(...)
        else: 
            # parameters of online spike stream from spike camera
            self.decode_width, self.height, self.width = ParseKwargs(...)
            
    def get_device_matrix(self, _input, _framepool, block_len=500):
        ... # get spike matrix from device (spike camera)
        
    def get_spikes(self, _input, _framepool):
        ... # reading Thread for online spike streams

    def get_spike_matrix(self, flipud=True):
        ... # get spike matrix from offline file
        
    def get_block_spikes(self, begin_idx, block_len=1, flipud=True):
        ... # get block spikes with begin index from file
        
    ... # Other Interfaces for Developers
\end{lstlisting}
\end{minipage}
\rule{0.47\textwidth}{1pt}
\caption{Definition of the class SpikeStream.}
\label{fig:spikeStream}
\end{figure}

There are three attributes of \textbf{SpikeStream} in default settings: the spike data variable \textit{SpikeMatrix}, the spatial resolution \textit{width}, and \textit{height}. As shown in Fig.~\ref{fig:spikeStream}, we offer two modes to initialize an instance of \textbf{SpikeStream}. The offline mode is used for loading spikes files. Users only need to deliver a parameter dictionary, including the file path and the resolution information. The function \textit{`get\_spike\_matrix'} and \textit{`get\_block\_spikes'} offer flexibility to load a full-length stream or a part of it. The online mode offers an interface to obtain real-time spikes from the spike camera. The function \textit{`get\_device\_matrix'} gets the handle of the device to control the process of reading data, and an instance of the caching frame pool. In this function, we apply a separate thread to parse spike streams to an array-like representation from the pool. The initialization step mainly unifies the online and offline modes of the loading process.

With the \textbf{SpikeStream}, fresh users do not need to bother implementing a new data-loading interface starting from scratch. Moreover, \textbf{SpikeStream} is compatible with the whole workflow in the SpikeCV, especially with the existing spike datasets, which brings convenience for developers to build and train a model fast.
Expandability is a necessity to encapsulate the spike data. For experienced users or the developer team, the object-oriented design benefits them by designing customized derived classes of the \textbf{SpikeStream}, adding new attributes, functions, or representations of spikes. In this way, users can encapsulate spikes freely and flexibly for more complex demands in projects.

\subsubsection{Standardization for Dataset Interfaces }
\label{Sec:standardization_for_dataset_interfaces}
We designed standard dataset interfaces in SpikeCV for various vision tasks for spike cameras. To support diverse tasks, we survey and collect the currently released dataset for spike cameras, whose detail is shown in Sec.~\ref{Subsection:Datasets}. The standard interfaces are compatible with various kinds of ground truths, including those for reconstruction, optical flow, monocular depth, stereo, detection, tracking, and recognition. Besides, the dataset interfaces in SpikeCV use standard parameter passing modes and support different file structures. Users can easily define dataset interfaces for their tasks based on the label interfaces, standard parameter passing, and standard \textbf{SpikeStream} class for reading spikes. 

\subsubsection{Modularization for Vision Tasks} 
The comprehensive modularization of SpikeCV enables different vision tasks, such as image reconstruction, optical flow, object detection, and depth estimation, to be easily implemented and integrated into the framework. This modular design allows developers to easily expand and customize the framework to fit their needs. To be specific, we have designed a series of utensils for different vision  tasks, such as spike pre-processing, data augmentation, dataloader, visualization and metric, which can be easily accessed and utilized by developers to improve the accuracy and assess the efficiency of different vision tasks.  Additionally, the model files are separated from the main code, providing an extra level of modularization and encapsulation.
SpikeCV provides a comprehensive and highly modularized solution for tackling a wide range of visual tasks, making it easy for developers to customize and improve algorithms. 

\subsection{Real-time}
Real-time is an important principle of SpikeCV to take advantage of the low latency of the spike streams. In order to achieve the online application, we carry out real-time processing from two aspects of data and reasoning. The detail of the real-time pipeline will be described in Sec.~\ref{sec:realtime_pipeline}.

\subsubsection{Real-time data}
To obtain real-time spike stream data from the spike camera, we design a multi-level thread for reading ultra-high temporal resolution spikes. Through the synchronous transmission of data streams between multiple C++ thread pools, spike data can be supplied to applications in real time without being blocked and lost.

\subsubsection{Real-time Inference}
Coupled with real-time data loading, we aim to achieve real-time inference and prediction. The encapsulated \textit{SpikeStream} instance can be produced from the pieces of real-time data from the spike camera, and be supplied to multiple parallel threads of different models. Each algorithm thread realizes the real-time operation of the spike vision application by simultaneously processing the latest data.

\begin{figure*}[ht]
    \centering
    \includegraphics[width=.9\linewidth]{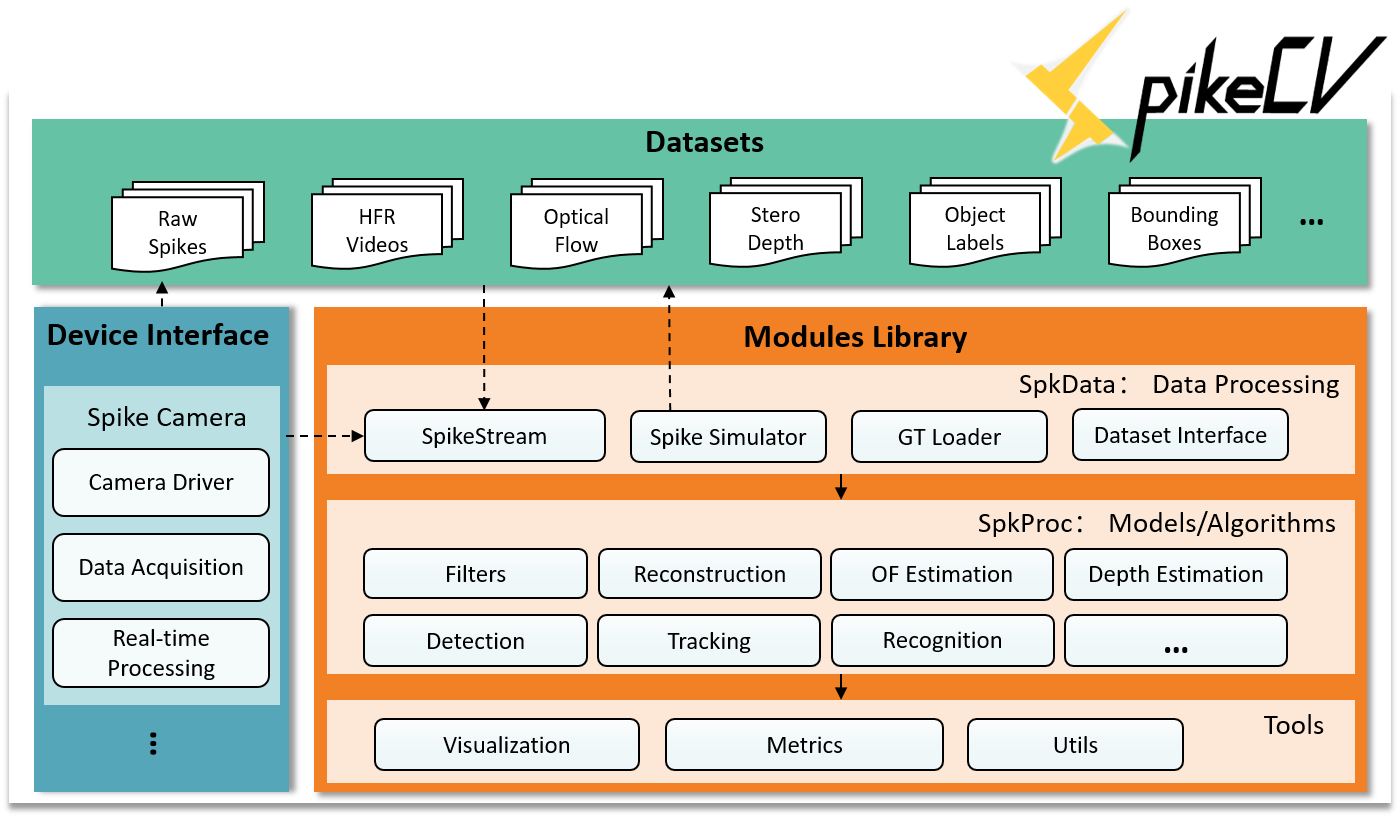}
    \caption{Illustration of the structure of the SpikeCV.}
    \label{fig:structure}
\end{figure*}

\subsection{Spike Ecosystem}
We aim to build up a comprehensive community ecology for the spike vision to facilitate the progress of scientific research as well as enterprise projects. For the community of image-based vision, many well-developed open-source libraries, like OpenCV~\cite{bradski2008learning}, benefit developers with various tools for image processing. As a novel type of vision sensor, spike cameras require an open platform to release available resources, which benefits researchers in getting familiar with spike vision and getting involved in practice quickly. Different from the design principles of existing image-based open-source libraries, SpikeCV not only provides spike processing tools and spike-based visual algorithms but also provides spike camera hardware interfaces and normalized spike datasets. In this highly integrated and modularized platform, beginners can thoroughly learn what spike data is and how to use spike cameras to tackle visual tasks. It helps draw more attention to establish the community and facilitate the ecology of spike vision.


\section{Details of SpikeCV}

\subsection{Architecture}

As illustrated in Fig.~\ref{fig:structure}, SpikeCV consists of three parts: spike datasets, modules libraries, and device interfaces.

\begin{itemize}
\setlength{\itemsep}{0pt}
\setlength{\parsep}{0pt}
\setlength{\parskip}{0pt}
    \item \textbf{Dataset:} Normalized datasets that can be used to validate or train intelligent models. The dataset files include the real scene generated by the spike camera or the spike simulator, and the corresponding label information such as video, optical flow, depth, and object category. We also provide supplementary configuration files for recording properties of spike data and label files.
    
    \item \textbf{Modules library:} The module library includes three categories: data processing, vision model/algorithm, and tools. The data processing module \textit{SpkData} includes the spike stream class, the spike simulator, the label loader, and the dataset interface for the training model. The algorithm library \textit{SpkProc} contains various vision algorithms for spike cameras, and the \textit{Tool} is a supporting development tool library responsible for visualization, metrics, and development utensils. 
    
    \item \textbf{Device interface:} The device interface is to facilitate the user to apply the spike-based algorithm to the real-time processing hardware. Currently, SpikeCV has integrated the spike camera. Users can use our spike camera interface to customize the scene to collect datasets or evaluate the real-time performance of the algorithm.
\end{itemize}


\begin{figure*}[th]
    \centering
    \includegraphics[width=0.98\linewidth]{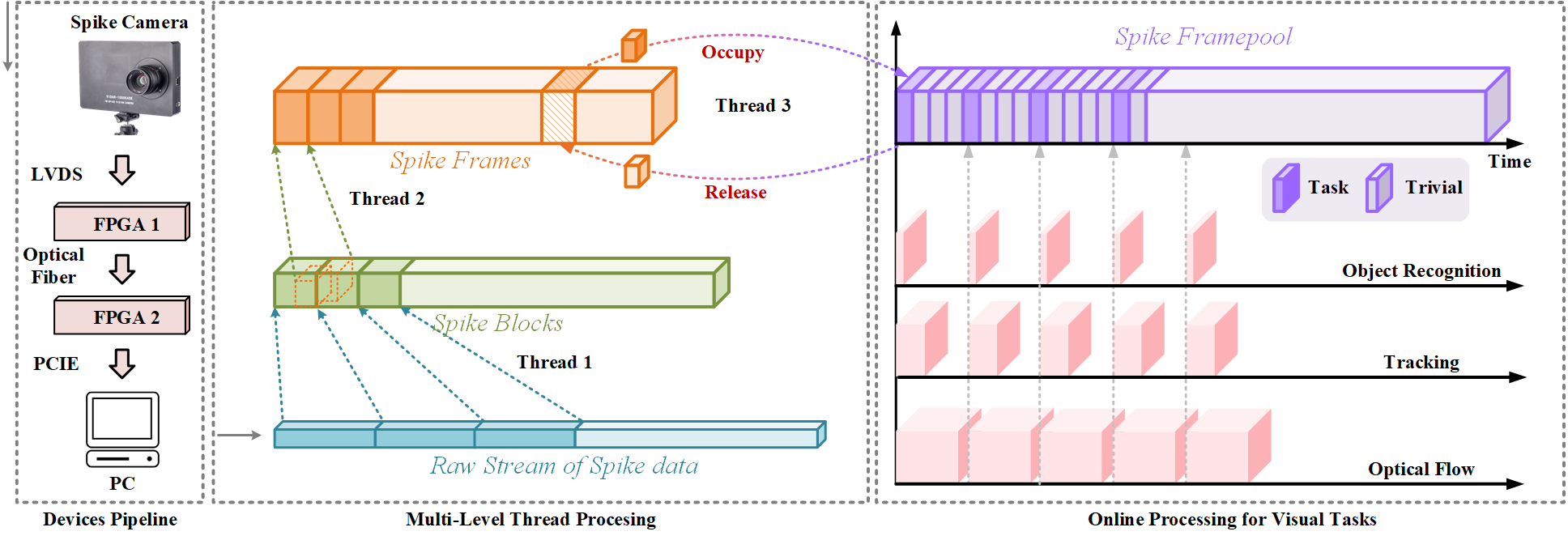}
    \caption{Illustration of the real-time pipeline of SpikeCV. The pipeline includes the strategy of parsing spike data with multi-level threading management and real-time processing for multi tasks.}
    \label{fig:realtime_pipeline}
\end{figure*}

Datasets and device interfaces are the two sources of data in SpikeCV for obtaining spike streams. These two sources of data have different purposes. The dataset mainly provides spike streams and corresponding labeled data that can be used for various visual tasks, which is convenient to evaluate the performance of spike-based algorithms or train machine learning models for different tasks. In addition to spike files, the dataset we provided also contains configuration files representing their spatial resolution, and label data properties. For data that has not been labeled, we use `raw' to indicate its label type. In addition to the datasets we provide, users can also use spike cameras through hardware interfaces to obtain real-time spikes to evaluate the performance of their models in various scenarios or build their datasets by customizing scenarios according to task requirements. Except for using the device interface to obtain real-world spike streams, users can also use the spike simulator in SpikeCV to convert high frame rate video or picture sequences into corresponding spikes, and quickly obtain labeled spike datasets.

\paragraph{Real-Time Pipeline}
\label{sec:realtime_pipeline}
We aim to synchronize the acquisition of data and the process of algorithms in a real-time scheme in the SpikeCV. As illustrated in Fig.~\ref{fig:realtime_pipeline}, we design a pipeline that can parse user-friendly spike frames from raw outputs of the hardware and process multiple visual tasks in real time. The pipeline mainly consists of two parts, multi-level threads for reading real-time spikes and online visual applications.

\renewcommand\arraystretch{1.5} 
\begin{table*}[ht]
    \centering
    \small
    \caption{Spike-based algorithm package in SpikeCV}
        \begin{tabular}{cccc}
            \toprule
            \textbf{Function}                & \textbf{Package name}             & \textbf{Input}                        & \textbf{Output}                                       \\ \hline
            Filters                 & spkProc.filters          & \multirow{8}{*}{SpikeStream} & Filtered spikes                            \\
            Reconstruction          & spkProc.reconstruction   &                              & Images/videos matrix                 \\
            Optical flow estimation & spkProc.optical\_flow     &                              & Optical Flow                           \\
            Depth Estimation        & spkProc.depth\_estimation &                              & Stero/monocular depth                   \\
            Object detection        & spkProc.detection        &                              & bbox and type                         \\
            Multi-object tracking   & spkProc.tracking         &                              & bbox and trajectory                            \\
            Object recognition      & spkProc.recognition      &                              & Object label                           \\
            Data augmentation       & spkProc.augment          &                              & Augmented spikes              \\ \bottomrule
        \end{tabular}%
    \label{tab:spkproc}
\end{table*}

\begin{figure}[htb]
    \rule{0.48\textwidth}{1pt}
    \begin{minipage}{0.48\textwidth}
\begin{lstlisting}
from spkProc.tracking.spike_sort import SpikeSORT
from visualization.get_video import obtain_detection_video, vis_trajectory

# MOT model construction
spike_tracker = SpikeSORT(spike_h, spike_w, device)
spike_tracker.calibrate_motion(spikes, calibration_time)
spike_tracker.get_results(spikes, tracking_file)
spike_tracker.save_trajectory(trajectories_filename)

#visualization
vis_trajectory(tracking_file, trajectories_filename, **para_dict)
obtain_mot_video(spikes, video_filename, tracking_file, **para_dict)

# measure performance
metrics = TrackingMetrics(tracking_file, **paraDict)
metrics.get_results()
\end{lstlisting}
    \end{minipage}
    \rule{0.48\textwidth}{1pt}
    \caption{Example of building a multi-object tracking model using SpikeCV}
    \label{fig:ssort}
\end{figure}

The first part is mainly implemented in the SpikeLinkInput Libary of SpikeCV with C++ language. Developers also can modify the C++ codes to meet different requirements. In this part, three threads are responsible for collecting spikes at different stages. \textbf{Thread 1} takes the raw stream of spikes and packs them into blocks that are arranged into a queue \textbf{LibBlockQueue}. The blocks contain spikes and other information like frame headers. \textbf{Thread 2} parses the spike blocks and assembles spikes into streams of bytes arranged in the spatial and temporal order. When the user enters a command to get camera data, these byte streams are continuously fetched and put into the queue \textbf{Lib-FramePool}.
\textbf{Thread 3} takes charge of handling spike frames between the C++ library and the user's APP in the Python program. In this thread, the byte streams are converted to pieces of user-friendly frames with the size of $H \times W \times T_{\text{cusum}}$, where $H \times W$ is the spatial resolution of the camera sensor and $T_{\text{cusum}}$ is a customized length of spike frames in one piece. 
 The APP also maintains a queue \textbf{App-FramePool}, and \textbf{Thread 3} always delivers the data address from \textit{Lib-FramePool} to \textit{App-FramePool}. After occupying the data, \textit{App-FramePool} will release the address back to \textit{Lib-FramePool}. 

The second part is the mechanism of simultaneously processing multiple tasks cooperated with the \textit{App-FramePool} in real time. Fig.~\ref{fig:realtime_pipeline} shows an example of three tasks processed by three independent threads, and the pink boxes denote the single-inference time cost. The main program always takes out frames from the \textit{App-FramePool}. One frame will be processed when three threads are all in idle states, otherwise, its pointer will be released right back. In this way, algorithms for different tasks always process the same frame and the \textit{App-FramePool} would not be blocked.

\paragraph{Spike Loader}
Data loading and processing is the key to building models. As described in Sec.~\ref{sec:encapsu_spikes}, we design SpikeStream as a data type for spikes to be compatible with different vision tasks and different data sources. Users can construct continuous spike stream data through SpikeStream, or convert it into a NumPy array, or tensor to perform matrix-based programming, and conveniently use the library algorithms supported by these data types.

For the data generated by the spike camera, SpikeStream will use a dedicated read thread to obtain the high-speed spike stream generated by the spike camera running in real time. These spike streams will be buffered into a memory pool by the input callback function of the hardware interface. The reading thread in SpikeStream reads from the memory pool at high speed, forms a temporally continuous spike matrix, and returns it for online processing. SpikeStream mainly represents the data type of the spike stream. To distinguish and unify the datasets corresponding to different vision tasks, we provide a data dictionary method \textit{`data\_parameter\_dict'} that specifically represents the attributes of the dataset. Users only need to specify the task type and the name of the dataset to use the obtained data dictionary to build the SpikeStream class or build a data loader for training machine learning models.

\paragraph{Model Construction}  

We provide various spike-based visual algorithms in SpikeCV, and users can choose different modules or combine them to construct the model according to task requirements. Tab.~\ref{tab:spkproc} introduces the Python package names, input, and output of the algorithm modules. These algorithms include adjusted traditional vision methods based on spikes (e.g., SVM~\cite{hearst1998support}), spiking neural networks~\cite{huang20221000x,zheng2021high}, and deep-learning models~\cite{chen2022self,zhang2022spike,zhao2022learning}.


\begin{figure*}[ht]
    \centering
    \includegraphics[width=\linewidth]{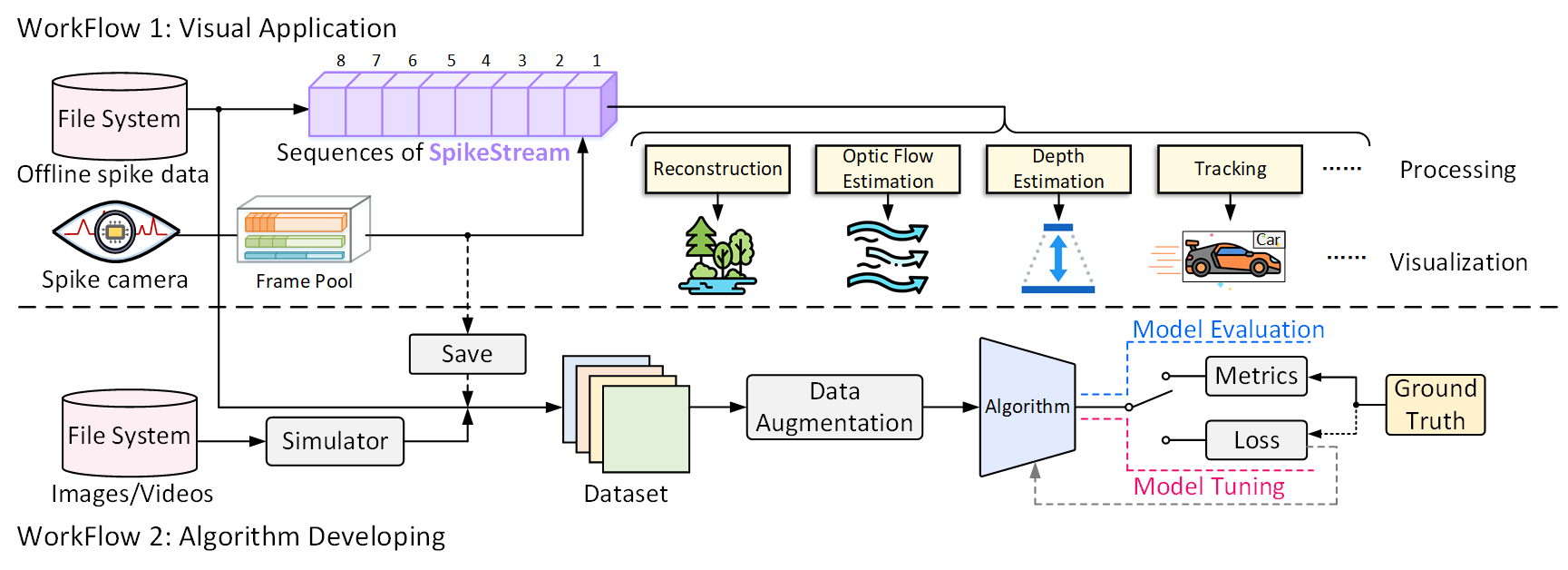}
    \caption{Illustration of the workflow of SpikeCV.}
    \label{fig:workflow_spikecv}
\end{figure*}

Reusability and configurability of models are the important properties of the modules for SpikeCV. When creating a model, users can train the model according to task goals or directly apply our vision algorithm. Taking the multi-object tracking task as an example, the currently integrated SNN-based algorithm~\cite{huang20221000x} is an online model, which does not require a large amount of data for model training. As shown in Fig.~\ref{fig:ssort}, developers can build the model with a few lines of code and directly apply the algorithm to track objects. With the obtained results, users can use the \textit{Visualization} package in SpikeCV to display the candidate boxes for detection and the trajectory of each object. If there is ground truth data for the corresponding task, the \textit{Metrics} module can be applied to quantitatively measure the performance of the models.

\subsection{Workflow}
The workflow of SpikeCV meets common needs of visual tasks and visualization. Fig.~\ref{fig:workflow_spikecv} shows the workflow of SpikeCV, which contains two parts of the visual application and algorithm development.
A typical SpikeCV workflow for developers is to conduct visual applications which involve data loading, processing, and visualization. Developers can utilize 
the data loading interface to load offline sequences of spike stream to the memory or collect real-time stream from real-time spike camera. The real-time stream can be either processed in the memory or saved to the file system for offline evaluation. The processing part manipulates the memory data and is responsible for high-quality reconstruction as well as performing various visual tasks. Developers can run vision tasks including optical flow estimation, depth estimation, object tracking, object detection, object recognition, etc. After processing the spike stream, developers can visualize the task output through the provided visualization module. These tools can generate continuous images and show the visualized results of different tasks such as series of optical flows and depth maps.

Furthermore, to allow developers to design better spike processing algorithms, we provide a wealth of tools to assist the development workflow. A high-quality dataset is essential for developers. We present three methods for obtaining datasets: one is to use pre-defined offline datasets, and the second is to simulate spike streams from images/videos using simulators. Besides, developers can also collect real-world data by saving real-time streams. After preparing the label of data, developers can use the encapsulated dataset modules to load the customized dataset and efficiently start model tuning and evaluation. Data argumentation can be performed to increase the generalization and robustness of models such as rotating, flipping, and resizing. To evaluate the models, developers can use pre-defined metrics and losses concerning ground truth to validate the model.

\subsection{Datasets}
\label{Subsection:Datasets}
SpikeCV offers detailed descriptions and interfaces for most of the currently released datasets for spike cameras. The details of the datasets are shown in Tab.~\ref{tab:datasets}. According to the ways to obtain the data, these datasets can be divided into real-captured and synthetic data. Some of the real-captured datasets have no labels and are mainly for subjective quality assessment for various tasks. The synthetic datasets are generated with different kinds of spike simulators. All of the synthetic datasets have image ground truths, and some of the synthetic datasets have ground truths of optical flow and depth maps. SpikeCV offers dataset interfaces that are easy to use, and an example of defining a dataset interface is shown in Procedure~\ref{procedure:using_dataset_interface}.

\floatname{algorithm}{Procedure}
\begin{algorithm}
\caption{Using Dataset Interface in SpikeCV} \label{procedure:using_dataset_interface}
\footnotesize
\begin{algorithmic}[1] 
\Statex {\textbf{/* Definition of the Dataset: */}}
\State {Initialize the parameters of the Dataset;}
\State {Construct the \small{\texttt{path\_list}} of the dataset;}
\State {Get samples based on the parameters and paths:}
\State {\quad (a) Construct \small{\texttt{data\_parameter\_dict}};}
\State {\quad (b) Construct \small{\texttt{SpikeStream}} object and other data;}
\State {\quad (c) Return the gotten data;}
\Statex {\textbf{/* Using of the Dataset: */}}
\State {Define parameters in \small{\texttt{data\_parameter\_dict}};}
\State {Create an object of the Dataset class based on the defined parameters;}
\State {Construct the loader using \small{\texttt{DataLoader}} in \small{\texttt{Pytorch}};}
\end{algorithmic}
\end{algorithm}

\renewcommand\arraystretch{1.5} 
\setlength\tabcolsep{7pt} 
\begin{table*}[htbp!]
  \centering
  \small
  \caption{Datasets about spike cameras that are open-source.}
  \begin{tabular}{ |c|c|c|l|c| }
    \hline
    & {Dataset} & {Label} & \makecell[c]{Description} & {Resolution} 
    \\ \hline
    \multirow{7}{*}{\rotatebox{90}{Real-captured}}
    & recVidarReal2019~\cite{zhu2020retina} & N/A & Scenes with high-speed objects or spike camera & $250 \times 400$ \\ \cline{2-5}
    & momVidarReal2021 & N/A & Various objects with different kinds of motion patterns & $250 \times 400$ \\ \cline{2-5}
    & PKU-Retina-Recon~\cite{zhu2019retina} & N/A & Spike and event data from the same scenes & $250 \times 400$ \\ \cline{2-5}
    & motVidarReal2020~\cite{zheng2022spike} & Tracking & Multi-object detection dataset for spike data & $250 \times 400$ \\ \cline{2-5}
    & PKU-Vidar-DVS~\cite{li2022retinomorphic} & Detection & Detection dataset for joint imaging with spike and event & $250 \times 400$ \\ \cline{2-5}
    & PKU-Spike-Stereo~\cite{wang2022learning} & Depth & Stereo depth dataset for spike data &  $250 \times 400$ \\ \cline{2-5}
    & Outdoor-Spike~\cite{zhang2022spike} & N/A & Spike data from autonomous driving scenes & $250 \times 400$ \\ \hline
    \multirow{7}{*}{\rotatebox{90}{Synthetic}}
    & Spike-REDS~\cite{zhao2021spk2imgnet} & Image & Synthesized from REDS dataset~\cite{Nah2019NTIRE} for reconstruction & $720 \times 1280$ \\ \cline{2-5}
    & Spike-Vimeo~\cite{xiang2021learning} & Image & Synthesized from Vimeo-90k dataset~\cite{xue2019video} for super-resolution & $256 \times 448$ \\ \cline{2-5}
    & SPIFT~\cite{hu2022optical} & Image, Flow & Simulated dataset with stochastic scenes for optical flow & $500 \times 800$ \\ \cline{2-5}
    & PHM~\cite{hu2022optical} & Image, Flow & Simulated dataset with designed scenes for optical flow &  $500 \times 800$ \\ \cline{2-5}
    & RSSF~\cite{zhao2022learning} & Image, Flow & Synthesized data
    for optical flow in real scenes & $\sim 0.6 \text{-} 1.3 \text{MP}$ \\ \cline{2-5}
    & Spike-KITTI~\cite{wang2022learning} & Image, Depth & Synthesized data from KITTI~\cite{menze2015object} for stereo depth & $\sim 370 \times 1240$ \\ \cline{2-5}
    & DENSE-Spike~\cite{zhang2022spike} & Image, Depth & Synthesized data from DENSE dataset~\cite{hidalgo2020learning} for monocular depth & $346 \times 260$  \\
    \hline
  \end{tabular}
  \label{tab:datasets}
\end{table*}

\section{Verification Applications}
In this section, we use the datasets, hardware interfaces, and module libraries in SpikeCV to implement offline inference tests and real-time applications for various vision tasks, respectively. All experiments are performed employing the SpikeCV module. 

\begin{figure}[htb]
    \centering
    \includegraphics[width=0.48\textwidth]{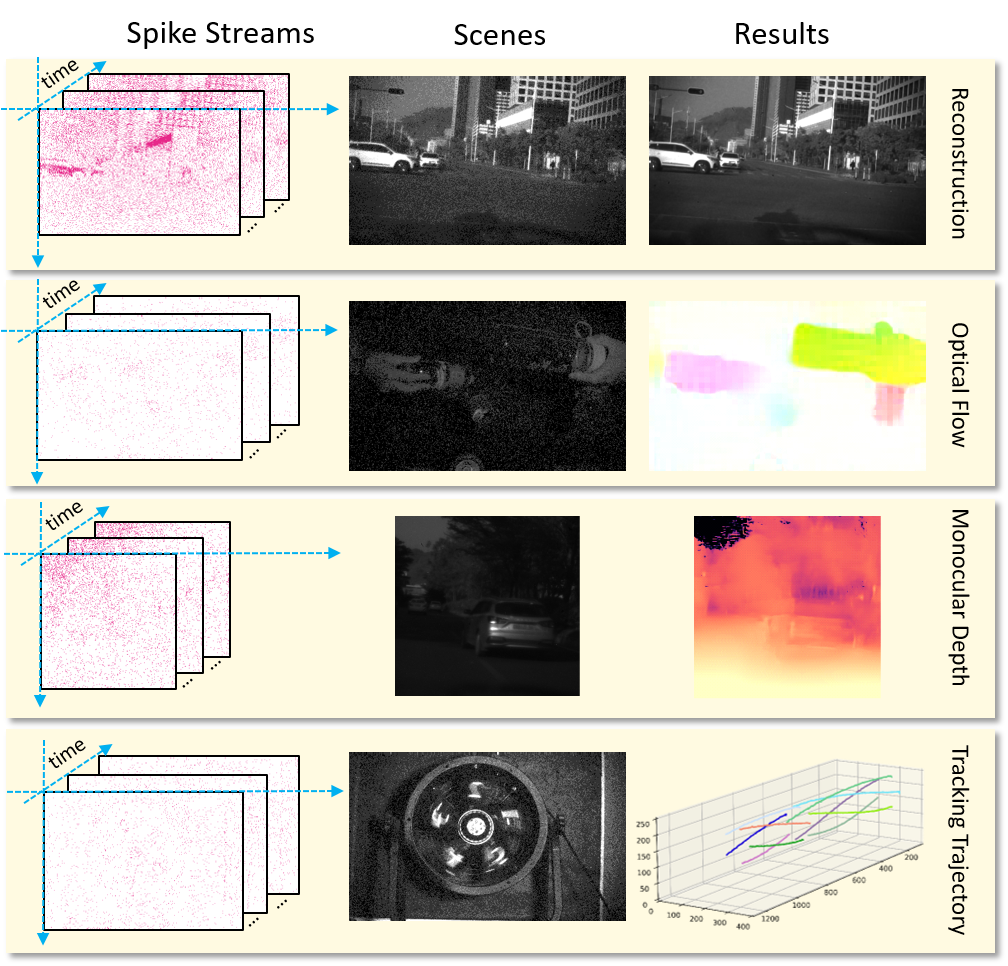}
    \caption{Examples of vision tasks implemented based on the offline dataset PKU-Vidar-DVS in SpikeCV. From top to bottom are image reconstruction, optical flow estimation, monocular depth estimation, and multi-object tracking tasks. Their input is the spike stream obtained by the data interface of SpikeCV. The scenes in the middle column are reconstructed with the TFP algorithm~\cite{zhu2019retina}, which has been brightened for visualization.}
    \label{fig:exp_offline}
\end{figure}

\subsection{Offline evaluation}

To verify the reusability of each set of algorithms in similar scenes, we use the PKU-Vidar-DVS dataset of SpikeCV that covers a variety of indoor and outdoor scenes and is rich in motion types for offline evaluation. We test the effects of image reconstruction, optical flow estimation~\cite{zhao2022learning}, monocular depth estimation~\cite{zhang2022spike}, and multi-object tracking algorithm~\cite{huang20221000x} modules on this dataset. Since the PKU-Vidar-DVS dataset~\cite{li2022retinomorphic} only provides label data applicable to object detection, we subjectively evaluate the results of these four tasks by visualizing the results. The visualization results of the experiments are shown in Fig.~\ref{fig:exp_offline}. It can be seen that the reproduction effects of these models in SpikeCV are promising. The first and third rows of Fig.~\ref{fig:exp_offline} show the outdoor autonomous driving scenario. In the driving scene, the SSML~\cite{chen2022self} reconstruction algorithm module can obtain images with less noise and sharp edges and contours, and the depth estimation module can generate depth maps with more details of contours. The second and last row of Fig.~\ref{fig:exp_offline} are scenes of multiple waving bottles and high-speed rotating characters shot indoors, respectively. The results of these two rows show that even in the case of sparse spike streams, the algorithm module can still robustly estimate the optical flow and trajectory of moving objects.

\subsection{Real-time applications}

\begin{figure}[htb]
    \centering
    \includegraphics[width=0.48\textwidth]{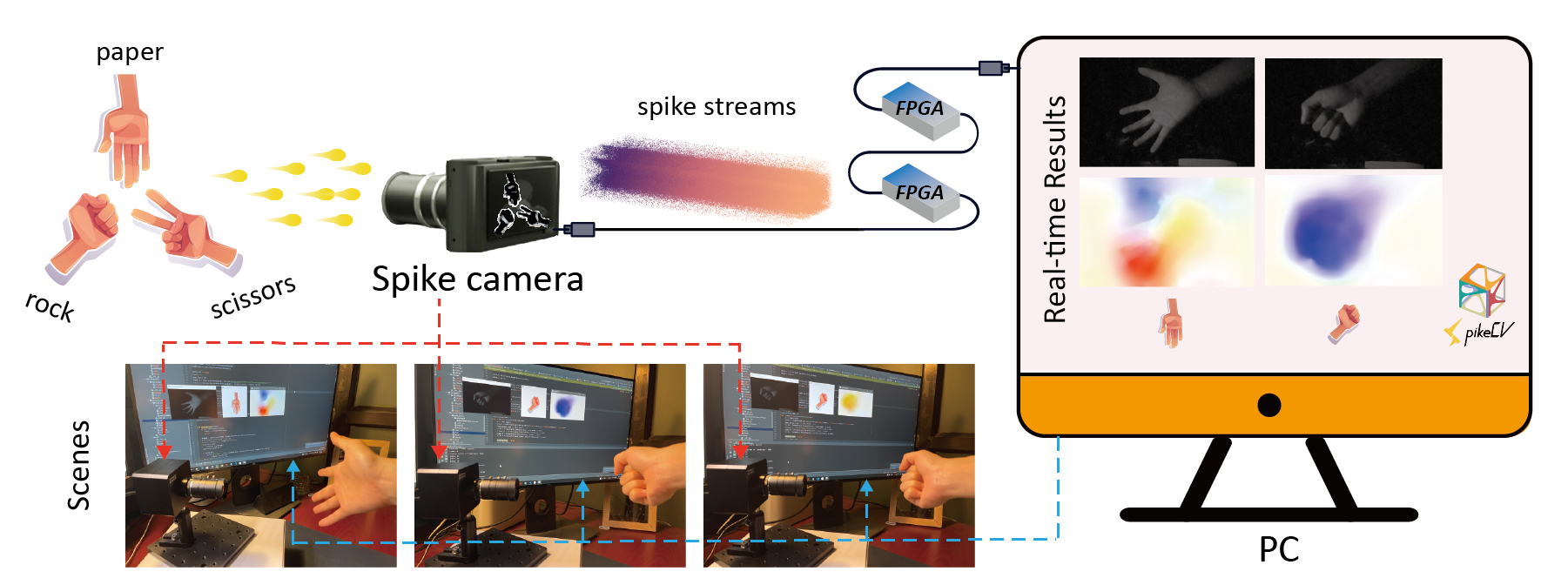}
    \caption{Real-time application scenarios of SpikeCV in rock-paper-scissors tasks}
    \label{fig:exp_online}
\end{figure}

In the game of rock-paper-scissors, only fast no broken. The game would be easy if seeing the opponent's gesture within a very small time interval. In the real-time application, we take advantage of the spike camera to collect continuous visual information and simultaneously apply the reconstruction, optical flow estimation, and recognition algorithms in SpikeCV to process the real-time spike streams generated by the camera, followed by using the visualization module to display the results interactively. Fig.~\ref{fig:exp_online} shows the running scenarios and results of the real-time application. As shown in the results, even with rapid changes of gestures (from paper to rock), the model constructed by SpikeCV can give high-quality results in time.

\section{Conclusion and future work}
In this paper, we introduce SpikeCV, an open-source platform for ultra-fast spike cameras, from aspects of design principles, workflow, and usage methods of each module in detail. At present, SpikeCV includes a rich set of normalized spike datasets, easy-to-use hardware interfaces, and a large library of spike-based algorithm modules. In the experiment, we demonstrate using SpikeCV to use offline datasets and online cameras as data sources and use each module in the algorithm library to achieve the tasks of image reconstruction, optical flow estimation, depth estimation, detection, tracking, and recognition. The experimental results show the effectiveness and usability of SpikeCV.

In the future, we will be committed to building a spike vision research community. We will provide large-scale spike datasets that can be applied to various algorithms, and continue to supplement and improve spike-based algorithms and processing tools. For the hardware interfaces, we plan to add more interfaces of cameras to facilitate the fusion of other data with the spike camera.
To take advantage of the spike camera as a neuromorphic sensor, we will provide the interface of the neuromorphic computing chip~\cite{akopyan2015truenorth,davies2018loihi,furber2016brain,pei2019towards}, to take advantage of low power consumption and ultra-high speed of brain-like vision algorithm~\cite{maass1997networks,neftci2019surrogate,tan2021strategy,wunderlich2021event}, and apply the spike vision to various ultra-fast scene applications, such as UAV~\cite{zhang2019slimyolov3} and automatic driving~\cite{muller2005off}.


{\small
\bibliographystyle{ieee_fullname}
\bibliography{egbib}
}

\end{document}